\documentclass[conference]{IEEEtran}
\IEEEoverridecommandlockouts
\usepackage[square,sort,comma,numbers]{natbib}
\usepackage[hidelinks]{hyperref}
\usepackage{url}

\usepackage{multirow}
\usepackage{amsmath,amssymb,amsfonts}
\usepackage[nameinlink]{cleveref}
\usepackage{algorithmic}
\usepackage{graphicx}
\usepackage{textcomp}
\usepackage[table]{xcolor}
\usepackage{makecell}
\usepackage{microtype}
\usepackage{xspace}
\usepackage{booktabs}
\usepackage{array}
\usepackage{tikz}
\usepackage{tcolorbox}
\usepackage{soul}
\usepackage{enumitem}
\usetikzlibrary{positioning, arrows.meta, shapes, shadows.blur, quotes}

\tikzset{
  block/.style={rectangle, rounded corners, text width=2cm, align=center, draw, blur shadow={shadow blur steps=5}, font=\small},
  connector/.style={-Stealth, thick}
}

\newcolumntype{P}[1]{>{\center asing\arraybackslash}p{#1}}
\newcolumntype{M}[1]{>{\centering\arraybackslash}m{#1}}

\def\BibTeX{{\rm B\kern-.05em{\sc i\kern-.025em b}\kern-.08em
    T\kern-.1667em\lower.7ex\hbox{E}\kern-.125emX}}

\makeatletter
\newcommand{\ie}{\emph{i.e.,~}\@ifnextchar.{\!\@gobble}{}}
\newcommand{\eg}{\emph{e.g.,~}\@ifnextchar.{\!\@gobble}{}}
\newcommand{\etc}{etc\@ifnextchar.{}{.\@}}
\makeatother

\definecolor{lightgray}{gray}{0.9}

\begin{document}


\title{Hardware, Algorithms, and Applications of the Neuromorphic Vision Sensor: a Review }

\author{\IEEEauthorblockN{Claudio Cimarelli$^{\ddagger}$, Jose Andres Millan-Romera$^{\dagger}$, Holger Voos$^{\dagger}$, and Jose Luis Sanchez-Lopez$^{\dagger}$}
\thanks{$^{\dagger}$Authors are with the Automation and Robotics Research Group, Interdisciplinary Centre for Security, Reliability, and Trust (SnT), University of Luxembourg, Luxembourg. Holger Voos is also associated with the Faculty of Science, Technology, and Medicine, University of Luxembourg, Luxembourg. \tt{\small{\{jose.millan, holger.voos, joseluis.sanchezlopez\}}@uni.lu}}
\thanks{$^{\ddagger}$Claudio Cimarelli was formerly with the University of Luxembourg and is now an independent researcher. Contact: \tt{\small{claudio.cimarelli@alumni.uni.lu}}} 
\thanks{*
This work was supported by the European Defence Agency (EDA) OB Study Neuromorphic Camera for Defence Applications
"Copyright @ - 2025 - European Defence Agency. All rights reserved.
The opinions expressed herein reflect the author's view only. Under no circumstances shall the European Defence Agency be held liable for any loss, damage, liability or expense incurred or suffered that is claimed to have resulted from the use of any of the information included herein."}%
}


\maketitle

\begin{abstract}

Neuromorphic, or event-based, cameras represent a transformation in the classical approach to visual sensing, encoding detected instantaneous per-pixel illumination changes into an asynchronous stream of event packets. Their novelty compared to standard cameras lies in the transition from capturing full picture frames at fixed time intervals to a sparse data format, which, with its distinctive qualities, offers potential improvements in various applications. However, these advantages come at the cost of reinventing algorithmic procedures or adapting them to effectively process the new data format.

In this survey, we systematically examine neuromorphic vision along three main dimensions. First, we highlight the technological evolution and distinctive hardware features of neuromorphic cameras from their inception to recent models. Second, we review image processing algorithms developed explicitly for event-based data, covering key works on feature detection, tracking, and optical flow, which form the basis for analyzing image elements and transformations, as well as depth and pose estimation or object recognition, which interpret more complex scene structures and components. These techniques, drawn from classical computer vision and modern data-driven approaches, are examined to illustrate the breadth of applications for event-based cameras. Third, we present practical application case studies demonstrating how event cameras have been successfully used across various industries and scenarios.

Finally, we analyze the challenges limiting widespread adoption, identify significant research gaps compared to standard imaging techniques, and outline promising future directions and opportunities that neuromorphic vision offers.

\end{abstract}




\section{Introduction}

Standard RGB cameras face considerable limitations, particularly in dynamic environments. The principle of these sensors involves capturing visual information as a sequence of frames at specific time intervals. Time-quantizing visual data at predetermined frame rates often results in temporal resolution limitations, as the frame rate is not aligned with the dynamic evolution of the scene. Consequently, significant details can be missed, especially in rapidly changing environments. Moreover, recording every pixel in each frame, regardless of changes since the last capture, leads to data redundancy, which affects data rate and volume~\cite{Posch2014}.

Instead, the limited dynamic range of standard RGB cameras often causes under- or overexposure in scenes with rapidly varying lighting conditions~\cite{Ceccarelli2023}. In addition, motion blur is another common problem in high-speed movement scenarios. Then, the latency inherent in fixed frame rate and power consumption for processing a large amount of data, \eg resulting from redundant information, poses an obstacle when real-time responsiveness and energy efficiency are required. 

In response to these limitations, neuromorphic cameras (NCs) or event cameras, as they are named more frequently in the robotics vision research domain, represent a paradigm shift~\cite{gallego2020event} in acquiring visual information compared to conventional frame-based cameras. In particular, each event camera's pixel operates independently \ie with its analog circuit, continuously comparing the current brightness to a reference level~\cite{Lichtsteiner2008}. When the difference exceeds a certain threshold, the pixel generates a sparse stream of event packets, \eg the pixel's address, timestamp, and the polarity of the brightness change, labeled with a high temporal resolution. This method allows for capturing visual information that mimics the human retina~\cite{Boahen}, approaching image acquisition with a biologically inspired process that responds more to real-world dynamics.

For their unique capabilities, NCs are highly suitable for various applications where real-time processing, adaptability to diverse lighting conditions, and energy efficiency are critical. These include robotics, surveillance, autonomous vehicles, and other areas that require robust and efficient visual sensing. For example, NCs can provide low-latency obstacle detection, even in challenging lighting or weather conditions, which is crucial in autonomous vehicle navigation. In robotics, event cameras enable more responsive situational awareness to changes in the dynamic environment. Due to its radically different sensor modality, NCs can offer non-invasive monitoring systems, which is helpful in data privacy-preserving scenarios like healthcare. Moreover, their low power consumption and small data volumes, \ie sparse event packets vs. dense image frames, make them ideal for remote surveillance monitoring systems or search and rescue missions, where energy efficiency is paramount. 

To date, neuromorphic vision technology and event camera image processing have been the focus of multiple reviews, each exploring the topic from a unique perspective. These surveys cover many aspects, from sensor technology to image processing methodologies. In \autoref{table:surveys} and \autoref{table:surveys2}, we overview their focus topic and highlight their key contributions.
The multitude of surveys reflects the rapid development of neuromorphic vision research in recent years and the traction the field has gained, especially in the robotics and computer vision communities. 

\begin{table*}[h!]
\centering
\resizebox{\linewidth}{!}{%
\begin{tabular}{M{2.5cm} M{1cm} M{2.8cm} M{8cm}}
\Xhline{5\arrayrulewidth}
\textbf{Reference} & \textbf{Year} & \textbf{Focus Topic} & \textbf{Key Highlights} \\
\Xhline{3\arrayrulewidth}
\rowcolor{lightgray}
\citet{adraEventbasedSolutionsHumancentered2025} & 2025 & Human-Centered Event-Based Applications & \vspace{3pt}\parbox{8cm}{Provides the first comprehensive survey unifying event-based vision applications for body and face analysis. Discusses challenges, opportunities, and less-explored topics such as event compression and simulation frameworks.}\vspace{3pt}\\

\citet{shariffEventCamerasAutomotive2024} & 2024 & Automotive Sensing (In-Cabin \& Out-of-Cabin) & \vspace{3pt}\parbox{8cm}{Presents a comprehensive review of event cameras for automotive sensing, covering both in-cabin (driver/passenger monitoring) and out-of-cabin (object detection, SLAM, obstacle avoidance). Details hardware architecture, data processing, datasets, noise filtering, sensor fusion, and transformer-based approaches.}\vspace{3pt}\\

\rowcolor{lightgray}
\citet{cazzatoApplicationDrivenSurveyEventBased2024a} & 2024 & Application-Driven Event-Based Vision & \vspace{3pt}\parbox{8cm}{Reviews event-based neuromorphic vision sensors from an application perspective. Categorizes computer vision problems by field and discusses each application area's key challenges, major achievements, and unique characteristics.}\vspace{3pt}\\  

\citet{chakravarthiRecentEventCamera2024} & 2024 & Event Camera Innovations & \vspace{3pt}\parbox{8cm}{Traces the evolution of event cameras, comparing them with traditional sensors. Reviews technological milestones, major camera models, datasets, and simulators while consolidating research resources for further innovation.}\vspace{3pt}\\  

\rowcolor{lightgray}
\citet{tenzin2024application} & 2024 & Event-Based VSLAM and Neuromorphic Computing & \vspace{3pt}\parbox{8cm}{Surveys the integration of event cameras and neuromorphic processors into VSLAM systems. Discusses feature extraction, motion estimation, and map reconstruction while highlighting energy efficiency, robustness, and real-time performance improvements.}\vspace{3pt}\\  

\citet{becattini2024neuromorphic} & 2024 & Face Analysis & \vspace{3pt}\parbox{8cm}{Examines novel applications such as expression and emotion recognition, face detection, identity verification, and gaze tracking for AR/VR, areas not previously covered by event cameras surveys. The paper emphasizes the significant gap in standardized datasets and benchmarks, stressing the importance of using real data over simulations.}\vspace{3pt}\\

\rowcolor{lightgray}
\citet{zheng2023deep} & 2023 & Deep Learning Approaches & \vspace{3pt}\parbox{8cm}{Extensively surveys deep learning approaches for event-based vision, focusing on advancements in data representation and processing techniques. It systematically categorizes and evaluates methods across multiple computer vision topics. The paper discusses the unique advantages of event cameras, particularly under challenging conditions, and suggests future directions for integrating deep learning to exploit these benefits further.}\vspace{3pt}\\

\citet{Huang2022} & 2023 & Self Localization and Mapping&\vspace{3pt}\parbox{8cm}{Discusses various event-based vSLAM methods, including feature-based, direct, motion-compensation, and deep learning approaches. Evaluates these methods on different benchmarks, underscoring their unique properties and advantages with respect to one another. Then, it gives deep reasons for the challenges inherent to sensors and the task of SLAM, drawing future directions for research.}\vspace{3pt}\\
 
\rowcolor{lightgray}
\citet{Shi2022}  & 2022 & Motion and Depth Estimation for Indoor Positioning & \vspace{3pt}\parbox{8cm}{Reviews notable techniques for ego-motion estimation, tracking, and depth estimation utilizing event-based sensing. Then, it suggests further research directions for real-world applications to indoor positioning.}\vspace{3pt}\\

\citet{Furmonas2022}  & 2022 & Depth Estimation Techniques &\vspace{3pt} \parbox{8cm}{Discusses various depth estimation approaches, including monocular and stereo methods, detailing the strengths and challenges of each. It advocates integrating these sensors with neuromorphic computing platforms to enhance depth perception accuracy and processing efficiency.}\vspace{3pt} \\

\rowcolor{lightgray}
\citet{Cho2022}  & 2022 & Material Innovations and Computing Paradigms  & \vspace{3pt}\parbox{8cm}{Highlights the evolution from traditional designs to innovative in-sensor and near-sensor computing that optimizes processing speed and energy efficiency. It addresses the challenge of complex manufacturing processes, suggesting directions for future research and application in flexible electronics.}\vspace{3pt}\\

\citet{Liao2021} & 2021 & Technologies and Biological Principles &  \vspace{3pt}\parbox{8cm}{Reviews advancements in neuromorphic vision sensors, contrasting silicon-based CMOS technologies such as DVS, DAVIS, and ATIS with emerging technologies in analogical devices. }\vspace{3pt}\\

\Xhline{5\arrayrulewidth}
\end{tabular}
}
\vspace{5pt} 
\caption{Overview of Previous Surveys on Neuromorphic Vision (Part 1)}
\label{table:surveys}
\end{table*}

\begin{table*}[h!]
\centering
\resizebox{\linewidth}{!}{%
\begin{tabular}{M{2.5cm} M{1cm} M{2.8cm} M{8cm}}
\Xhline{5\arrayrulewidth}
\textbf{Reference} & \textbf{Year} & \textbf{Focus Topic} & \textbf{Key Highlights} \\
\Xhline{3\arrayrulewidth}
\rowcolor{lightgray}

 \citet{gallego2020event} & 2020 & Sensor Working Principle and Vision Algorithms & \vspace{3pt} \parbox{8cm}{Thoroughly reviews the advancements in event-based vision, emphasizing its unique properties. The survey spans various vision tasks, including feature detection, optical flow, and object recognition, and discusses innovative processing techniques. It also outlines significant challenges and future opportunities in this rapidly evolving field.}\vspace{3pt}\\

\citet{Steffen2019} & 2019 & Stereo Vision and Sensor Principles & \vspace{3pt}\parbox{8cm}{Performs a comparative analysis of event-based sensors, focusing on technologies such as DVS, DAVIS, and ATIS. It reviews the biological principles underlying depth perception and explores the approaches to stereoscopy using event-based sensors.}\vspace{3pt}\\

\rowcolor{lightgray}
\citet{Lakshmi2019} & 2019 & Object Motion \& SLAM & \vspace{3pt}\parbox{8cm}{ Reviews state-of-the-art event-based vision algorithms for object detection/recognition, object tracking, localization, and mapping. Highlights the necessity of adapting conventional vision algorithms. Also, provides an overview of publicly available event datasets and their applications.} \vspace{3pt}\\

\citet{Vanarse2016} & 2016 & Neuromorphic Vision, Auditory, and Olfactory Sensors &\vspace{3pt} \parbox{8cm}{Highlights low power consumption in the prototypical developments of DVS and DAVIS using asynchronous spiking output. Suggests future research directions in neuro-biological emulating sensors for vision, audition, and olfaction with multi-sensor integration.}\vspace{3pt}\\
\Xhline{5\arrayrulewidth}
\end{tabular}}
\vspace{5pt}
\caption{Overview of Previous Surveys on Neuromorphic Vision (Part 2)}
\label{table:surveys2}
\end{table*}

We aim to bridge the gap between technological advancements in neuromorphic vision and their practical adoption across industries. To achieve this, we provide a comprehensive overview of the evolution of neuromorphic sensors, detailing their functionalities, algorithmic developments, and real-world applications. By outlining recent research and developments, we identify and highlight the technical and practical limitations of current event-based vision by comparing it with standard image sensors and addressing the challenges of adapting classical algorithms. More importantly, we also uncover opportunities for future advancements and broader adoption of event-based vision systems, emphasizing their unique advantages across various applications. 

In summary, with this survey, we make several distinct contributions:
\begin{itemize}
    \item \textbf{Neuromorphic Cameras' Hardware Evolution}: We present a timeline of the evolution of neuromorphic vision sensor technology, revealing the chronological progress of the hardware, how it differs from standard vision systems, and why these differences matter.
    
    \item \textbf{Event-based Image Processing \& Algorithms}: We examine the progression of image-processing techniques from classical methods to advanced deep-learning approaches.
    
    \item \textbf{Application Focus}: We discuss key application case studies demonstrating how the unique properties of neuromorphic cameras impact real-world solutions. 
    
    \item \textbf{Gaps, Limitations, and Future Opportunities}: We analyze the key challenges hindering the adoption of neuromorphic vision sensors, from hardware constraints to algorithmic gaps and real-world application barriers, while highlighting the opportunities unlocked by this radical shift in visual sensing modality.
\end{itemize}

\section{The Neuromorphic Vision Sensor}
The initial concept of NC invention arrived from the research group of Professor Carver Mead at Caltech and with the publication of the book "Analog VLSI and Neural Systems" in 1989~\cite{Mead1989}. 
Notably, Misha Mahowald, Mead's student, developed during her Ph.D. from 1986 to 1992 at Caltech the first neuromorphic chip to spike events resulting from detected light intensity variations~\cite{Mahowald1991, Mahowald1991a, Mahowald1992}. 
The first specialized commercial application inspired by these ideas was a motion detection system for pointing devices designed by Xavier Arreguit of CSEM for Logitech in 1996~\cite{Arreguit}. 

The following discussion clarifies the pixel design and its asynchronous data output, illustrates camera models that came to the market from visible light to the infrared spectrum, and concludes by examining the unique characteristics of event cameras that make them a distinctive technology.


\subsection{The Neuromorphic Camera's Asynchronous Photoreceptor}

Neuromorphic cameras leverage asynchronous photoreceptors to efficiently mimic the responsiveness and energy efficiency of the human visual system, a concept that has been explored since the early developments of the silicon retina \cite{Mahowald1991}. These receptors detect changes in light intensity and encode this information into discrete events, resulting in an array of pixels operating independently. This mechanism contrasts with the conventional camera's approach of capturing entire frames regularly, thus processing and transmitting large volumes of redundant data.

Central to the operation of neuromorphic cameras is the Asynchronous Address-Event Representation (AER), an innovative communication protocol developed from the pioneering research by the Caltech group led by Carver Mead~\cite{Mahowald1991a} and refined through subsequent research \cite{Boahen2000}. AER uses time-coded addresses to encode and dynamically transmit events between the silicon photoreceptor and the computing processor. Each event, whether an ON-event indicating an increase in light or an OFF-event indicating a decrease, is defined by its pixel reference, timestamp, and polarity.

\begin{figure}[!h]
    \centering    \includegraphics[width=\linewidth]{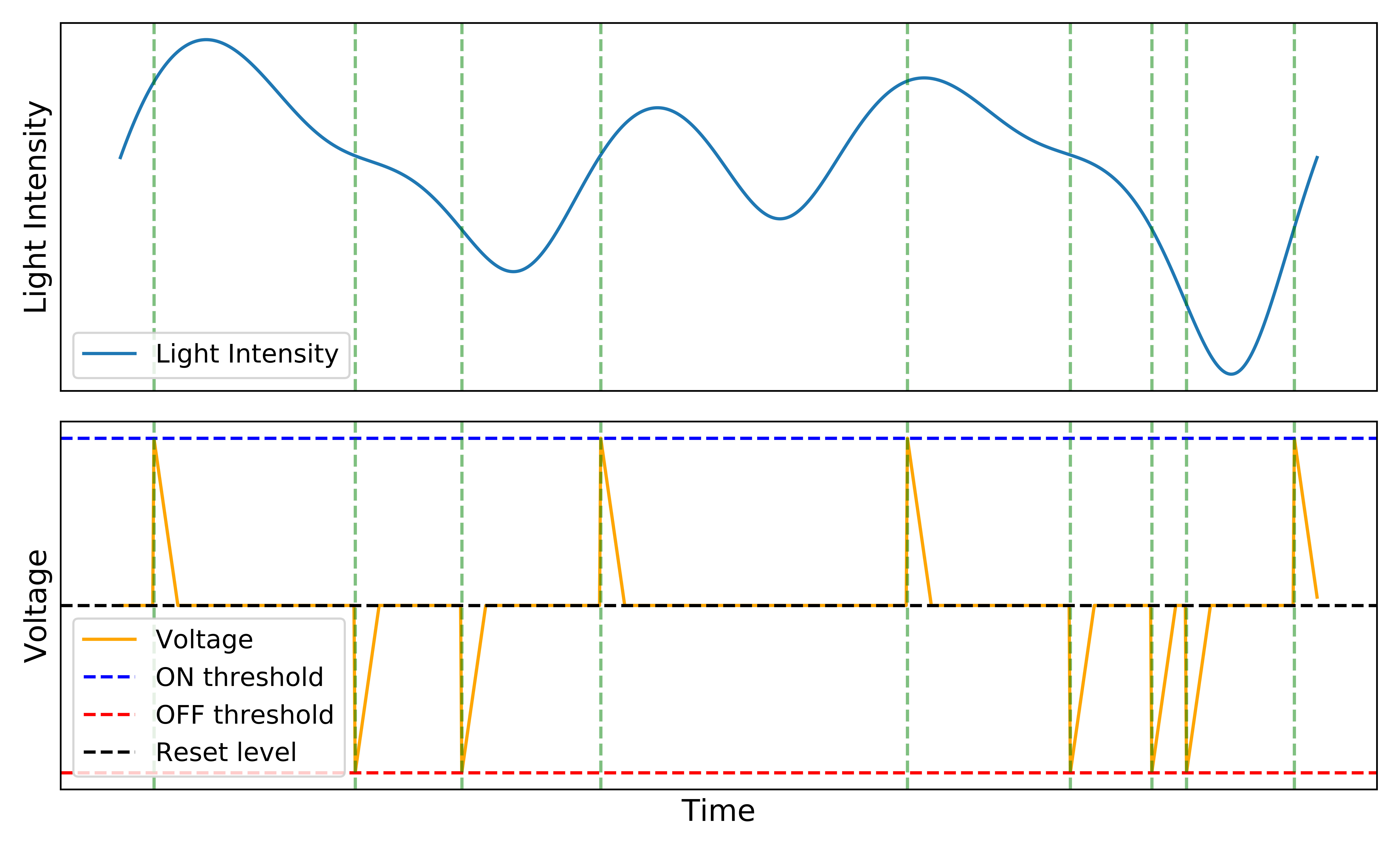}
    \caption{The asynchronous output of an operating event camera photodiode. Inspired from image in~\cite{Lichtsteiner2008}}
    \label{fig:log_photo}
\end{figure}

The neuromorphic photoreceptors respond to light intensity variations on a logarithmic scale. This capability allows the sensor to handle various lighting conditions effectively, from dim to bright. Each pixel analog circuit, as the primary designed in \cite{lichtsteiner2006128}, detects changes that surpass a voltage threshold, encoded in the photoreceptor, triggering the transmission of the AE packet, as illustrated in \autoref{fig:log_photo}. 

As explored by \citet{Steffen2019}, this sophisticated protocol incorporates a digital bus system and multiplexing strategies that allow all pixels to transmit their information over the same line efficiently and asynchronously, significantly reducing power consumption and data volume.

Further refining the process, the AER's implementation via address encoders generates unique binary addresses for each pixel event. This overall strategy highlights the role of AER in transmitting only essential visual information while discarding irrelevant static scenes and ensuring effective responses to rapid changes in the environment~\cite{Lazzaro1995}.

\subsection{Progresses of Visible-Light Event Camera's Models}

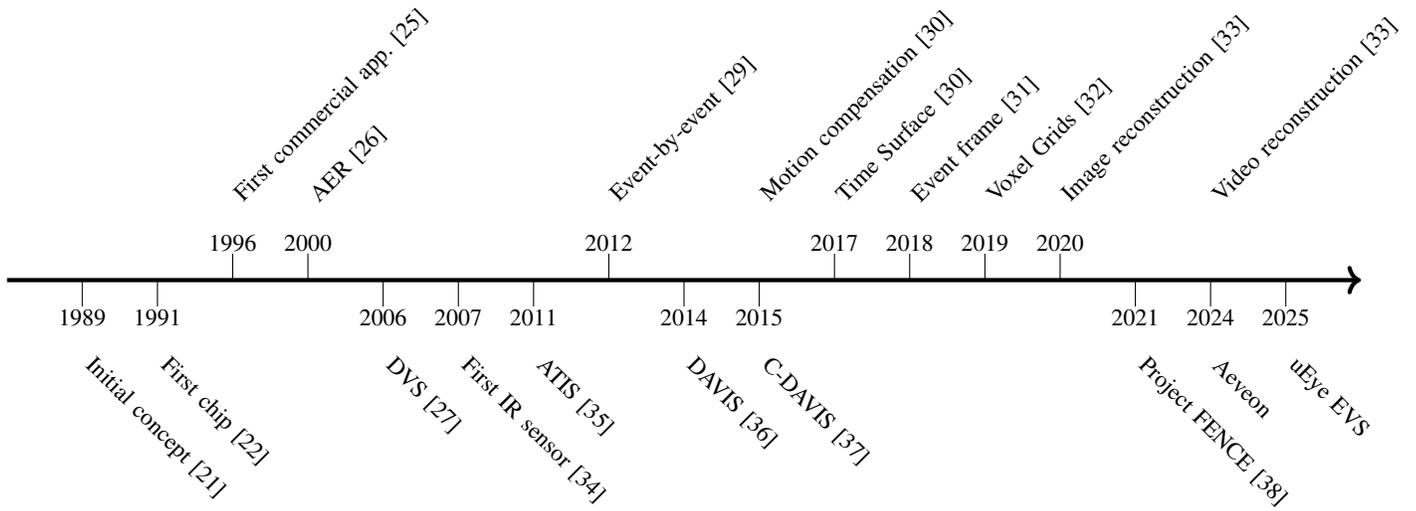
\begin{figure*}[htbp]
    \centering
\begin{tikzpicture}
    \draw[ultra thick, ->] (-1,0) -- (17, 0);


    \foreach \x in {2., 3, 7, 10,11, 12, 13} {
        \draw (\x cm, 0pt) -- (\x cm, 10pt);
    }

    \node[rotate=45, anchor=west, font=\small] at (2, 1) {First commercial app.~\cite{Arreguit}};    
    \node[font=\small] at (2, 0.5) {1996};
    
     \node[rotate=45, anchor=west, font=\small] at (3., 1) {AER~\cite{Boahen2000}};
    \node[font=\small] at (3, 0.5) {2000};
    
    \node[rotate=45, anchor=west, font=\small] at (7, 1) {Event-by-event~\cite{Weikersdorfer2012}};
    \node[font=\small] at (7, 0.5) {2012};

    \node[rotate=45, anchor=west, font=\small] at (9, 1) {Motion compensation~\cite{Lagorce2017}};
    
    \node[rotate=45, anchor=west, font=\small] at (10, 1) {Time Surface~\cite{Lagorce2017}};
    \node[font=\small] at (10, 0.5) {2017};

    \node[rotate=45, anchor=west, font=\small] at (11, 1) {Event frame~\cite{Liu2018}};
    \node[font=\small] at (11, 0.5) {2018};
    
    \node[rotate=45, anchor=west, font=\small] at (12, 1) {Voxel Grids~\cite{Zhu2019}};
    \node[font=\small] at (12, 0.5) {2019};
    \node[rotate=45, anchor=west, font=\small] at (13, 1) {Image reconstruction~\cite{Wang2020}};
    \node[font=\small] at (13, 0.5) {2020};
    
    \node[rotate=45, anchor=west, font=\small] at (15, 1) {Video reconstruction [33]};
    
    \foreach \x in {0., 1, 4, 5, 6, 8, 9, 14, 15, 16} {
        \draw (\x cm, -10pt) -- (\x cm, 0pt);
    }
      \node[rotate=-45, anchor=west, font=\small] at (0, -1) {Initial concept~\cite{Mead1989}};
    \node[font=\small] at (0, -0.5) {1989};    
    
    \node[rotate=-45, anchor=west, font=\small] at (1, -1) {First chip~\cite{Mahowald1991}};
    \node[font=\small] at (1, -0.5) {1991};
    \node[rotate=-45, anchor=west, font=\small] at (4, -1) {DVS~\cite{lichtsteiner2006128}};
    \node[font=\small] at (4, -0.5) {2006};
     \node[rotate=-45, anchor=west, font=\small] at (5, -1) {First IR sensor~\cite{Posch2007}};
    \node[font=\small] at (5, -0.5) {2007};
    \node[rotate=-45, anchor=west, font=\small] at (6, -1) {ATIS~\cite{Posch2011}};
    \node[font=\small] at (6, -0.5) {2011};
    \node[rotate=-45, anchor=west, font=\small] at (8., -1) {DAVIS~\cite{Brandli2014}};
    \node[font=\small] at (8, -0.5) {2014};
    \node[rotate=-45, anchor=west, font=\small] at (9, -1) {C-DAVIS~\cite{li2015design}};
    \node[font=\small] at (9, -0.5) {2015};

    \node[rotate=-45, anchor=west, font=\small] at (14, -1) {Project FENCE~\cite{darpa}};
    \node[font=\small] at (14, -0.5) {2021};
    
    \node[rotate=-45, anchor=west, font=\small] at (15, -1) {Aeveon};
    \node[font=\small] at (15, -0.5) {2024};
    
    \node[rotate=-45, anchor=west, font=\small] at (16, -1) {uEye EVS};
    \node[font=\small] at (16, -0.5) {2025};
    
    
\end{tikzpicture}
\caption{Timeline of pivotal milestones in developing the neuromorphic vision sensor. Events related to software developments are depicted above the axis, whereas hardware-related milestones are positioned below it.}
\label{fig:timeline}
\end{figure*}

From the theoretical progress in the 90s, several research organizations have started producing the first neuromorphic sensors. Toby Delbruck proposed the first generic event camera in 2008 in collaboration with Patrick Lichtsteiner and Christoph Posch under Dynamic Vision Sensor (DVS), the earliest event camera technology. Lichtsteiner, Posch, and Delbruck~\cite{lichtsteiner2006128, Lichtsteiner2008} proposed a novel silicon retina design that outputs AER in a 128$\times$128 pixel grid. Since the DVS's inception, many companies have commercialized VGA to megapixel resolution event cameras, from a small reality like CelePixel or Insightness to a technological giant like Samsung,  producing further innovations of the original sensor. Event camera expression gradually took place in the last few years to highlight the AER output of neuromorphic vision devices and differentiate them from their standard camera counterparts.

In 2014, Christoph Posch co-founded Chronocam (now Prophesee) in France, focusing on the development and commercialization of the Asynchronous Time-based Image Sensor (ATIS) technology,  outlined in Posch's 2011 research at the Austrian Institute of Technology~\cite{Posch2011}. ATIS marks a significant advancement in event camera technology, merging the temporal contrast-detection capabilities of the DVS with innovative time-based intensity measurement pixels. This integration allows ATIS to capture event-based data and provide absolute brightness measurements with high accuracy. However, incorporating a pulse width modulated (PWM) intensity readout mechanism for each DVS pixel, aimed at enhancing reconstruction and recognition capabilities, necessitated an extra photodiode per pixel, effectively doubling the pixel size. Moreover, because the PWM readout process required the transfer of triple the data amount, the ATIS latency is significantly increased, particularly impacting the sensor's ability to capture fast-moving or dimly lit objects. Despite this, the sensor's QVGA resolution (\eg 304$\times$240) substantially improves detail and image quality over the original DVS.
Furthermore, the ATIS addresses critical limitations of traditional imaging systems by significantly reducing temporal redundancy and delivering a high dynamic range (143 dB static and 125 dB at 30 FPS). Further developments of the ATIS technology have led to multiple sensor generations, such as Prophesee Metavision GEN3, including collaboration with Sony on the IMX636 and IMX637 sensors. These sensors, featuring stacked CMOS technology, underscore ATIS's ongoing evolution and potential in various high-performance imaging applications.

The Dynamic and Active-pixel Vision Sensor (DAVIS)~\cite{Brandli2014}, developed by IniVation, represents a significant advancement in vision sensor technology. Unlike its predecessors, DAVIS integrates neuromorphic event-driven and active pixel sensors (APS) functionality within the same photodiode. This innovative design enables the DAVIS to interleave event data with conventional intensity frames, using a shared pixel to generate grayscale and event data. The pixel architecture of DAVIS offers several benefits: it achieves a dynamic range of 130 dB for event detection and 51 dB for grayscale intensity frames. Additionally, it features a minimized latency of just 3 $\mu$s. Despite its dual functionality, the pixel area in DAVIS is only marginally larger (about 5\%) than that of a standard DVS pixel, resulting in a slightly reduced high dynamic range compared to the ATIS but in a more compact form factor.

The Color Dynamic and Active-Pixel Vision Sensor (C-DAVIS) represents a significant advancement, building upon the foundations of the earlier DAVIS  model~\cite{li2015design}. This sensor combines monochrome event-based pixels with a 5-transistor APS architecture integrated under a Red, Green, Blue, and White (RGBW) color filter array. Capable of outputting both rolling or global shutter RGBW-coded VGA resolution frames and asynchronous monochrome QVGA temporal contrast events, C-DAVIS excels in capturing vibrant color details as well as tracking swift movements with remarkable temporal precision. This blend of capabilities is efficiently packed into a compact design, featuring a 2$\times$2-pixel RGBW unit with dimensions of merely 20$\mu$m $\times$ 20$\mu$m, showcasing C-DAVIS's ability to combine high-resolution color imaging with fast, event-based motion detection.

In 2023, IniVation introduced the Aeveon sensor, an advancement in neuromorphic vision technologies, to address the limitations of previous models like the DAVIS. The Aeveon is designed to allow each pixel to generate several event types, including full pixel value (RGB), multi-bit and single-bit change events, and area events. Moreover, it employs a stacked sensor design with Adaptive Event Cores, merging characteristics of neuromorphic sensors with frame-based sensors. This design is compatible with various pixel types, from standard RGB to infrared. Furthermore, the sensor offers the flexibility to select an adaptable region of interest (ROI), where the user could focus the event stream reception, similar to an attention mechanism. With its unified solution, Aeveon should facilitate the integration with existing systems while providing an immediate replacement for the current vision modules and a pathway to introduce new event-based features gradually.

\begin{table*}[!htpb]
\centering
\resizebox{\textwidth}{!}{%
\begin{tabular}{ >{\centering}p{2.9cm} >{\centering}p{2.8cm} >{\centering}p{2cm} >{\centering}p{1.8cm} >{\centering}p{1.5cm} >{\centering}p{2.8cm} >{\centering}p{2cm} >{\centering}p{1.3cm} >{\centering\arraybackslash}p{1.8cm} }
\Xhline{3\arrayrulewidth}
\textbf{Manufacturer} & \textbf{Model} & \textbf{Resolution} & \textbf{Latency   } & \textbf{Temporal Resolution} & \textbf{Max Throughput} & \textbf{Dynamic Range} & \textbf{Power} & \textbf{Image Frames} \\
\Xhline{3\arrayrulewidth}
IniVation & DAVIS346 (also, Color) & 346 $\times$ 260 &  $<$ 1ms & $1\mu$s & 12 Meps & 120 dB & $<$ 180 mA& Graysc. / Color \\
\rowcolor{lightgray}
IniVation & DVXplorer  & 640 $\times$ 480 & $<$ 1ms & 65 - 200$\mu $s & 165 Meps & 90 - 110 dB & $<$ 140 mA & No \\
IniVation & DVXplorer Lite & 320 $\times$ 320 & $<$ 1ms & 65 - 200$\mu $s & 100 Meps & 90 - 110 dB & $<$ 140 mA & No \\
\rowcolor{lightgray}
IniVation & DVXplorer Micro & 640 $\times$ 480 & $<$ 1ms & 65 - 200$\mu $s & 450 Meps & 90 - 110 dB & $<$ 140 mA & No \\
Prophesee & Gen 3 VGA CD & 640 $\times$ 480 & 40 – 200 $\mu$s & NA & 66 Meps & $>$ 120 dB & NA &  No \\
\rowcolor{lightgray}
Prophesee & GENX320 & 320 $\times$ 320 &   $<$ 150 $\mu$s &  1 $\mu$s & NA &  $>$ 120 dB &  $ > $ 36 $\mu$W  & No\\
Sony / Prophesee & IMX636  & 1280 $\times$ 720 & 100 - 220 $\mu$s & NA & 1060 Meps & 86 dB & NA & Grayscale \\
\rowcolor{lightgray}
Sony / Prophesee & IMX637  & 640 $\times$ 512 & 100 - 220 $\mu$s & NA & 1060 Meps & 86 dB & NA & Grayscale \\
Sony / Prophesee &  IMX646 & 1280 $\times$ 720 & 800 - 9000 $\mu$s & NA & 1060 Meps & 110 dB & NA & Grayscale \\
\rowcolor{lightgray}
Sony / Prophesee &  IMX647 & 640 $\times$ 512 & 800 - 9000 $\mu$s & NA & 1060 Meps & 110 dB & NA & Grayscale \\
Imago Tech. / Prophesee & Vision Cam EB & 640 $\times$ 480 & 200 $\mu$s & NA & 30 Meps & $>$ 120 dB & NA & No \\
IDS / Sony / Prophesee & uEye EVS & 1280 $\times$ 720 & $<$ 100 $\mu$s  & $<$ 100 $\mu$s &  & $>$ 120 dB & 10 $\mu$W & No \\
\Xhline{3\arrayrulewidth}
\end{tabular}
}
\vspace{5pt} 
\caption{Comparison of currently commercially available event cameras. Meps is millions of events per second.}
\label{table:event_cameras}
\end{table*}

In \autoref{table:event_cameras}, we provide a list of currently commercialized event cameras. Other sensors have been listed in the literature, such as in~\cite{gallego2020event,Furmonas2022}. However, not all are available for purchase, \eg the early DVS128 and DVS240 from IniVation, or not easy to procure online, as the models produced by Insightness, Samsung~\cite{Son2017}, and CelePixel~\cite{Chen2012, Chen2019}. Recently, IDS and Prophesee partnered to create the new uEye EVS camera series~\cite{IDSProphesee2025}, enabling ultra-fast imaging (sub-100$\mu$s resolution) and significantly reducing data processing and power consumption.

Notably, the price of such devices is still a few thousand dollars, making them currently functional only for industrial purposes. The cost of production is the main obstacle to the diffusion of event cameras in larger commercial markets until mass production of the sensor's silicon. However, recent partnerships, such as the announced collaboration between Prophesee and Qualcomm, allow us to foresee that event cameras may soon be adopted for mobile platform imaging. Google is adding event-based vision to its Visual Intelligence and Android XR platforms—paving the way for advanced AR glasses~\cite{wired2024androidXR}. SynSense has launched the Speck neuromorphic vision SoC for ultra-low-power, high-speed imaging~\cite{synsense2022speck}.

\subsection{Development of the Infrared Neuromorphic Vision Sensor}
In the field of neuromorphic chip research, the primary emphasis has been on the visible waveband. However, there are certain situations when objects of interest become challenging to perceive due to fluctuations in the scene illumination. This problem becomes more pronounced when the photons of interest are not emitted in the visible waveband, such as during nighttime or when the atmospheric conditions are unsuitable for the visible waveband. 

To this aim, one approach is to shift or extend the measurable light spectrum toward the infrared region~\cite{boettiger2020comparative}. These sensors are typically categorized based on the wavelength range they are sensitive to, which includes short-wave infrared (SWIR), mid-wave infrared (MWIR), and long-wave infrared (LWIR). Each sensor type has its advantages and is suitable for different applications. SWIR cameras typically operate in the wavelength range of 1$\mu$m to 2.5$\mu$m. SWIR can distinguish between organic and inorganic materials, making them ideal for the recycling and food industry, agriculture using drones to detect lack of water, detect the lasers used in the military domain, or benefit from a better atmospheric transmission. MWIR cameras usually use a wavelength range of 3 to 5 $\mu$m. MWIR sensors are known for their ability to detect thermal radiation emitted by objects at high temperatures, making them suitable for airborne and ground-based surveillance, thermography, and gas detection applications. LWIR cameras typically operate in the wavelength range of 8 to 14 $\mu$m, commonly referred to as the thermal imaging region, as it allows the detection of the thermal radiation emitted by objects or materials at ambient temperature. LWIR sensors suit thermal imaging, night vision, and medical diagnostics applications.

\citet{Posch2007} developed the first IR event-based sensor by coupling a microbolometer array with typical DVS readout circuitry~\cite{Posch2009}. A microbolometer is a thermal sensor that detects thermal infrared radiation based on the variation of its temperature-dependent electrical resistance, and it is sensitive in the LWIR range. It can be integrated with complementary metal-oxide semiconductor (CMOS) readout circuitry. However, the time constant of current microbolometers (around 10 ms) is relatively slow and does not allow us to take full benefit of the NC technology. Alternative IR technologies, such as cryogenic IR quantum sensors for MWIR and LWIR or InGaAs for the SWIR region, seem more promising.

Furthermore, SCD is announcing the preparation of a product in SWIR~\cite{Jakobson2022}. This product with a resolution of VGA (15$\mu$m pitch) for the imaging mode and quarter-VGA for the event-based output should be available shortly under the name SWIFT-EI. In addition, DARPA has launched the FENCE project~\cite{darpa} to develop event-based infrared cameras sensitive to the infrared band above 3$\mu$m, supposedly including MWIR and LWIR.


\begin{table}[!htpb]
\centering
\resizebox{\linewidth}{!}{%
\begin{tabular}{ >{\centering}p{1.4cm} >{\centering}p{1.6cm} >{\centering}p{2.3cm} >{\centering\arraybackslash}p{3cm} }
\Xhline{3\arrayrulewidth}
\textbf{Category} & \textbf{Wavelength} & \textbf{Capability} & \textbf{Applications} \\
\Xhline{3\arrayrulewidth}
Short-wave (SWIR) & [1$\mu$m - 2.5$\mu$] & Organic vs. Inorganic & Recycling, food, agriculture and military\\
\rowcolor{lightgray}
Mid-wave (MWIR) & [3$\mu$m - 5$\mu$] & Thermal radiation & Surveillance, thermography and gas detection \\
Long-wave (LWIR)& [8$\mu$m - 14$\mu$] & Thermal radiation \& ambient temperature & Imaging, night vision and medical diagnosis \\
\Xhline{3\arrayrulewidth}
\end{tabular}
}
\vspace{5pt} 
\caption{Comparison of infrared neuromorphic cameras by wavelength.}
\label{table:infrared}

\end{table}

\subsection{Main Characteristics of Event Cameras}

The unique design and operational characteristics of neuromorphic vision sensors offer several advantages over traditional vision technologies. Here, we list the most remarkable:

\begin{itemize}
    \item \textbf{High temporal resolution}: NCs can capture fast-moving objects and obtain greater detail of the evolution of the motion without having to interpolate between frames. Light intensity change is detected by analog circuits with high-speed response. Then, a digital read-out with a 1MHz clock timestamps the event with microsecond resolution~\cite{gallego2020event}.
    \item \textbf{Low latency}: Event cameras have low latency, meaning they can respond quickly to environmental changes. Contrary to the traditional camera, NCs do not have a shutter, so there is no exposure time to wait before transmitting the brightness change events. Therefore, latency is often tens to hundreds of microseconds for laboratory conditions to a few milliseconds for real conditions~\cite{gallego2020event}. 
    \item \textbf{High dynamic range}: NCs can capture bright and dark scenes without losing detail. This property is particularly beneficial in sudden changes of illuminations that can cause overexposure or in low-light environments where the scene may appear too dark. This property is due to the logarithmic response at the photoreceptors. Hence, whereas static vision sensors have a limitation to the dynamic range at 60 dB because all the pixels share the same measurement integration time (dictated by the shutter), event cameras can go over 120 dB for their independent pixel operations.
    \item \textbf{Low power}: NCs consume much less energy than traditional cameras, making them suitable for low-power devices and applications. Notably, all pixels are activated independently based on the illumination changes each one detects, and the analog circuit is very efficient. As a result, the NCs' power demand can go as low as a few milliwatts.
    \item\textbf{Sparsity}: NCs provide data only when there is a change in the scene. Hence, the amount of data that needs to be processed is reduced.  As a result, the event cameras may output up to 100x less data than traditional cameras with similar resolution. To fully exploit this technology, NCs must be coupled with chips capable of processing the events with algorithms such as Spiking Neural Networks designed to maintain the low power premises and keep events' intrinsic asynchronous nature intact. SynSense is one of such companies that develops neuromorphic chips, such as the DYNAP-CNN chip, for ultra-low-power applications, \eg IoT devices, and can be integrated with the same chip with the DVS pixel array, as demonstrated by IniVation Speck.
\end{itemize}

\section{Working with Stream of Events}

The intrinsically different nature of the neuromorphic sensor compared to traditional cameras necessitates new approaches to represent the information captured and to process it in a format suitable for input into specific image-processing algorithms. Neuromorphic cameras (NCs), also known as event cameras, trigger asynchronous events for those pixels that detect a brightness change exceeding a certain threshold—caused either by camera motion or by moving objects in the scene. As a result, the sensor outputs so-called events that encode not only the spatial location of the pixel ($x$ and $y$ coordinates) but also the polarity of the brightness change (positive or negative) and a precise timestamp.

\begin{figure}[htpb] \centering \includegraphics[width=\linewidth]{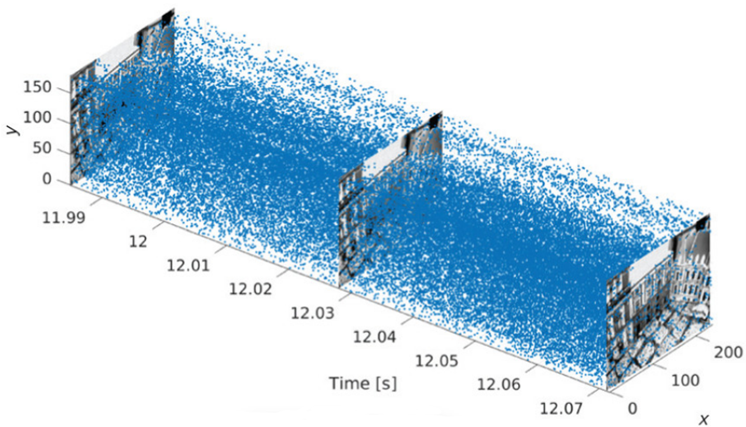} \caption{The stream of events as blue points interleaved with grayscale image frames at a fixed rate, as provided by a DAVIS event camera. Image credits to~\cite{Mueggler2017}.} \label{fig:event_stream} \end{figure}

An example of such an event stream is shown in~\autoref{fig:event_stream}, captured alongside fixed-rate grayscale frames. This visual juxtaposition highlights the drastic difference from the classic image input format.

This fundamentally different data format calls for new analysis methods~\cite{gallego2020event}. One direction is to develop algorithms that operate directly on the sparse and asynchronous event stream. Alternatively, the event stream can be converted into more conventional representations, which require fewer or no modifications to existing algorithms. Another design choice is whether to process individual events to minimize latency or to group events into packets, which can then be transformed into other formats and offer more contextual information. In either case, prior context must be considered, as a single event alone lacks sufficient information~\cite{Alevi2022}.

\subsection{Event-by-Event Processing}

Among the methods that process event-by-event, we recognize probabilistic Bayesian filters, \eg Kalman Filters (KF) or Particle Filters (PF) and Spiking Neural Networks (SNN). Bayesian filtering is a statistical method that relies on Bayesian theory to maintain a probability distribution over the possible states and update this distribution as new data becomes available. Examples of this approach are found especially concerning pose estimation~\cite{Weikersdorfer2012, Censi2014}. \citet{Gallego2018} demonstrate the tracking of the 6-DoF pose of a DVS camera from an existing photometric depth map. \citet{Kim2014} shed accurate rotational motion tracking while reconstructing high dynamic range spherical mosaic views from gradient mages using Poisson solvers~\cite{Tumblin2005}. \cite{Scheerlinck2018} introduced a continuous-time formulation for intensity estimation and fusion of events with image frames using a complementary filter. Additionally, the paper provides a new dataset for evaluating image reconstruction. Later, they proposed a method to compute the spatial convolution of a linear kernel with the output of an event camera, using an internal state that encodes the convolved image information and demonstrates the application of the proposed method to Harris corner detection~\cite{Scheerlinck2019}.

Instead, the Spiking Neural Network (SNN)~\cite{Lee2016} is a type of neural network that models the behavior of biological neurons and the communication between them using discrete spike signals. Hence, they can process asynchronous inputs by encoding the spike's timing. Typical use of these networks comprehends character recognition as demonstrated in~\cite{Orchard2015}, where an SNN architecture named HFirst exploits the event's temporal information, integrating Integrate-and-Fire neurons with a Winner-Take-All selection strategy. While in HFirst, the network comprises handcrafted kernels such as the Gabor filter~\cite{granlund1978search}, the SLAYER algorithm~\cite{Shrestha2018} demonstrates how to handle the non-differentiable nature of the spike signal. It performs a modified backpropagation to learn the weights and axonal delay parameters of SNNs. The SNN asynchronous and sparse spiking pattern can be exploited by specific neuromorphic hardware, such as the Intel Loihi~\cite{Davies2018}, to achieve highly power-efficient models compared to the traditional Deep Neural Networks (DNNs) running on GPUs. Unfortunately, this type of hardware is not yet commercially available, so we have to rely on conventional chips on which SNNs do not have the same energy efficiency properties. Therefore, converting packets of events into 2D or 3D representations is often convenient to process by computer vision algorithms that can better use the currently available hardware resources. In the following, we will explore the most common ways of pre-processing event packets and transform them into a format that allows standard image processing algorithms to analyze them. 

\subsection{Event Frame}

One of the first 2D representations of event packets is the Event Frame, which helps process event streams using traditional computer vision techniques, algorithms, and tools not explicitly designed for event-based data. Furthermore, the event frame representation can be used to visualize the events in a way familiar to human observers. In this representation, the events accumulate over time and are used to update a brightness increment image. The advantage of event frames is that their frame rate can be adapted to the use case. However, they have severe limitations compared to other representations, such as time surfaces or voxel grids, in capturing the temporal dynamics of the events. Examples of event frame applications include optical flow~\cite{Liu2018}, stereo vision~\cite{Kogler2009}, and deep learning applied to steering angle prediction~\cite{Maqueda2018}. 

\subsection{Time Surface}
Another popular representation is the Time Surface (TS)~\cite{Lagorce2017, Sironi2018} or Surface of Active Events (SAE)~\cite{Benosman2014}. A TS is a spatio-temporal representation of an event and its surrounding activities that use the arrival time of events from nearby pixels. It is a 2D array where each pixel stores the time of the most recent event at that location, with the pixel's intensity indicating the event's time. Time surfaces are a time-resolved version of an image and can be used to analyze the dynamics of an event stream over time. Recent events are emphasized over past events using an exponential kernel, and normalization is used to achieve invariance to motion speed. Each pixel value can be computed by filtering events within a space-time window to reduce the sensitivity to noise. However, time surfaces compress information by keeping only one timestamp per pixel, which can reduce their effectiveness on scenes with frequent events or textures.

\subsection{Voxel Grids and Point Sets}
Voxel grids involve dividing a 3D space into a regular grid of voxels, essentially 3D pixels, associated with a value representing the features or characteristics of the object or scene at that location. For example, each voxel contains the number of events within a spatio-temporal volume in the event camera context. The temporal dimension is discretized in multiple bins, and the voxel value is found by bilinear interpolation~\cite{Zhu2019}. Voxel grids help represent volumetric data in a structured way that neural networks can process efficiently. 

A similar approach represents events directly as 3D point sets, where each point is associated with an event fired at a particular time. For example,~\citet{Benosman2014} employ this representation to estimate motion velocity as a vector proportional to the slope of a plane fitted on the set of points. 

\subsection{Motion Compensation}
Motion compensation~\cite{Gallego2018} is a technique that represents events as image frames to reduce the motion blur and visualize sharp edges. It involves accumulating events over a certain period and using them to update an image representation of the scene that considers the motion. The intuition is that event cameras capture how edges move in the scene and are used to align the events that trigger them. Hence, we optimize an objective function called the focus function~\cite{Gallego2019} to find the trajectories that warp the event back to a reference time to maximize the visual sharpness. As a result, the resulting image is sharper, making it more informative and interpretable than a raw event stream. Hence, the feature extraction~\cite{Gehrig2020} and visual odometry~\cite{Rebecq2017a} tasks are easier to approach using the produced sharp edge map. In addition, motion compensation can be used with other event representations, such as time surfaces~\cite{Zhu2017} or 3D point sets~\cite{Mitrokhin2018}. 

\subsection{Image Reconstruction}
Image reconstruction involves obtaining grayscale frames of the scene from accumulated events. If standard camera images are available, they can be fused to add more information and overcome visual defects such as motion blur and limited dynamic range.
Some of the most relevant image reconstruction techniques for these types of cameras include:
\begin{itemize}
    \item\textbf{Spike-based image reconstruction}~\cite{Zhu2021} involves accumulating spike data over time and reconstructing an image of the scene. One common approach is to use a spike-based reconstruction algorithm that considers the spatial and temporal patterns of the spikes to recreate a picture that closely approximates the original scene.
    \item\textbf{Adaptive filtering}~\cite{Wang2020} filters out noise and artifacts in data captured by event cameras. Because these cameras capture data asynchronously and at high temporal resolution, there is often a lot of noise in the data that can interfere with image reconstruction. Adaptive filtering techniques use a combination of statistical analysis and machine learning algorithms to filter out the noise and improve the quality of the reconstructed image.
    \item\textbf{Compressed sensing}~\cite{Duan2021} can reconstruct high-quality images from relatively small quantity data using a combination of algorithms and mathematical models.
    \item\textbf{Deep learning}~\cite{Han2020} methods entail using neural networks to learn patterns in visual data and generate high-quality reconstructed images. This technique involves training a neural network on a large dataset of visual data and using it to reconstruct images from the sparse data captured by NCs and event cameras. Deep learning has shown promising results in improving the quality of reconstructed images from these types of cameras.
\end{itemize}


\subsection{Learning-Based Representations}
Finally, it is possible to create novel grid-based or memory-driven representations by end-to-end learning with neural networks. \citet{Annamalai2021} introduce a deep learning memory surface, which encodes temporal motion history directly from sparse events. Designed for anomaly detection, this representation preserves the asynchronous nature of the data while enabling efficient spatiotemporal analysis. Building on this, \citet{Teng2022} propose Neural Event Stacks (NEST), a novel spatiotemporal encoding that respects physical constraints while effectively capturing motion dynamics. Their learned representation achieves state-of-the-art performance on image enhancement tasks such as deblurring and super-resolution.

\citet{Gehrig2019} propose Event Spike Tensors (EST), a representation optimized for learning-based pipelines. The authors also introduce a taxonomy of event representations, distinguishing between hand-crafted and learned formats. \citet{Vemprala2021Representation} use event variational autoencoders (VAEs) to handle environmental changes effectively. \citet{Schaefer2022AEGNN} process event data as evolving spatio-temporal graphs, named AEGNN. Unlike previous methods that convert event streams into dense representations, AEGNN treats events as sparse data, updating only relevant parts of the graph.

\citet{Guo2023Compact} address the efficient representation of volumetric videos with feature grids and introduce dynamic codebooks for storage optimization. \citet{Wang2024EAS-SNN} develop an adaptive sampling approach that dynamically selects the most relevant events in the input stream. They also introduce EAS-SNN, a spiking neural network (SNN), to enhance temporal learning by using recurrent connections that preserve context over time. \citet{Gu2024Learning} improve event-based video reconstruction by learning contrast-threshold-adaptive parameter representations, addressing issues like blurry outputs and artifacts.



\section{Image Processing Algorithms}

\subsection{Extraction and Tracking of Image Features}
Identifying distinctive and informative features in visual data is the first step to further analyze and understand the surrounding world through the eyes of the camera. Therefore, extracting significant features is critical to distill sensorial input into condensed information that more complex algorithms can use. In practice, features enable visual tasks for a higher level of comprehension or situational awareness~\cite{bavleSLAMSituationalAwareness2023}, like object recognition, image retrieval, camera localization, or 3D reconstruction. In traditional image analysis, features are extracted from pixel intensity patterns corresponding to geometric structures like corners and edges. Classical methods such as Harris~\cite{Harris1988}, HOG~\cite{dalal2005histograms}, FAST~\cite{Rosten2006}, SIFT~\cite{lowe2004distinctive}, SURF~\cite{bay2006surf}, and ORB~\cite{rublee2011orb} enable feature detection and description, ensuring robustness to transformations in scale, rotation, and illumination. These methods rely on dense image frames, where feature vectors encode the appearance of local pixel neighborhoods for matching and tracking.

In contrast, event-based vision requires adapting feature extraction methodologies due to its sparse and asynchronous nature. This necessitates feature detection techniques that leverage the temporal structure of event streams rather than relying on fixed-frame representations.
~\citet{Vasco2016} propose an adaptation of Harris, while~\cite{Mueggler2017a} advanced a version of the FAST corner detection developed to work on time-surface representations of event streams. Instead,~\citet{Clady2015} find corners as the intersection of planes fitted on the time surface.~\citet{Alzugaray2018} presented an efficient version of eFAST for asynchronous corner detection called Arc. Subsequently, they build the ACE tracker~\cite{Alzugaray2018a} that uses a normalized local region descriptor applied to corners. FA-Harris~\cite{Li2019} is a faster corner detection method inspired by the Harris detector. To achieve speed, they introduce a Global Surface of Active Events (G-SAE) unit and corner candidate selection and determine detection scores, showing improved accuracy performance.~\citet{Li2021} move towards more complex descriptors constructed using the gradient information from Speed Invariant Time Surfaces (SITS)~\cite{Manderscheid2019}. DART~\cite{Ramesh2020} uses a log-polar grid to obtain a robust descriptor valuable for object detection and tracking. Recently, deep learning-based descriptors have started to appear.~\citet{Huang2022} propose a variation of the TS representation, Tencode, that considers polarities and creates a multi-temporal resolution input for training a deep network inspired by the Superpoint architecture~\cite{DeTone2018}. Their approach, EventPoint, shows promising results concerning previous methods, \eg \cite{Chiberre2021}, where Harris corners are extracted from predicted image gradients instead.

Extraction and tracking are intertwined tasks as features are good if we can track them for long frame sequences~\cite{Shi1994}. Feature tracking refers to establishing correspondences or matches between visual features over time in a sequence of images or video frames through a process usually referred to as data association. Feature tracking is typically performed by detecting the key points in the first frame of the sequence and then matching them with key points in subsequent frames. Matching can be done using various techniques, such as nearest-neighbor matching. The objective of tracking is to obtain a model of the motion between frames of the visual features, usually obtained by minimizing an objective function, such as reprojection or photometric error functions. 
 
Matching features detected in consecutive frames is usually done using Iterative Closest Point (ICP)~\cite{Besl1992} as in~\cite{Ni2012}, where large polygonal shapes are tracked. Notably, also~\citet{Tedaldi2016} and similarly~\citet{Kueng2016} track with ICP binary templates obtained with the Canny edge detection algorithm~\cite{Canny1986} centered around Harris corners. Instead, in previous approaches, the model template is generated from predefined patterns. For example, complex-shaped objects tracked by gradient descent~\cite{Ni2015} or by multiple kernels, \eg~Gabor filters, feeding a Gaussian tracker~\cite{Lagorce2015}. \citet{Glover2017} use a particle filter to improve over the previous approach applying Hough transform to track a fast-moving ball~\cite{Glover2016}.

~\citet{Zhu2017a} approach the data association problem with a probabilistic framework that jointly optimizes the matching with the feature displacements in an Expectation-Maximization scheme.~\citet{Gehrig2020} propose EKLT that resolves the data association challenge with a generative model to predict the future appearance of generic features. Hence, they use a Maximum Likelihood Estimation (MLE) to optimize the warp parameters and brightness increment velocity. While most previous techniques operate on intermediate representations that accumulate events in a traditional frame format, HASTE~\cite{Alzugaray2020} aims to track on an event-by-event basis. Hence, they revisit a previous tracker formulation~\cite{Alzugaray2019} with an efficient evaluation of the alignment score function that determines the transition among a discretized space of hypothetical states. Event Clustering-based Detection and Tracking (eCDT)~\cite{Hu2022} solves detection and tracking simultaneously with a novel clustering method that separates event groups based on the neighboring polarity and spatiotemporal adjacency. Finally,~\cite{Messikommer2022} train a neural network to predict displacements employing a correlation layer.

\subsection{Optical Flow}

Optical flow~\cite{Horn1981} is a technique for estimating the motion of objects or scenes in a sequence of images or video frames based on the apparent movement of pixels between frames. Optical flow computes a dense vector field that represents the displacement of each pixel in the image over time. Also, it can be used to track objects or scenes by identifying regions of similar optical flow using clustering techniques, such as mean shift. The task of optical flow is closely related to feature tracking, differing to compute a displacement vector for every pixel in the input frame rather than sparse keypoints regardless of the detection algorithm. However, due to incomplete information,~\eg the lack of knowledge of the scene geometry, ambiguity, noise, and occlusion, the problem is ill-defined and requires additional constraints. For example, brightness constancy assumptions~\cite{Barron1994} or local smoothness prior~\cite{Lucas1981} are usually applied.

Unlike image frames, events do not contain the same amount of information that can be extracted from observing the absolute brightness directly on an image plane. Hence, early methods, such as~\cite{Benosman2012, Benosman2014, Brosch2015, Orchard2013} start with testing optical flow reconstruction on the simple motion vector field created by a rotating black bar pattern, which triggers events on a continuous spiral in the $x-y-t$ space. \citet{Benosman2012} propose an algorithm based on the LucasKanade~\cite{Lucas1981} coarse-to-fine iterative approach by computing partial derivative over a small neighborhood of events. Later,~\citet{Benosman2014} refine this approach with an alternative formulation that finds the flow as the slope of a plane fitted on a spatiotemporal region of the event stream. \cite{Brosch2015} make multiple considerations on the previous approaches, such as the numerical instability of the gradient approximation approach or plane fitting that requires now too small nor too many events for robust estimation. Hence, they suggest a methodology that measures velocity as the response to a family of Gabor filters to different velocities and directions by fitting their frequency sensitivity on the experimental data~\cite{Valois2000}. These early approaches have been compared on a common benchmark where the optical flow was generated from a camera rotating on its three axes, and an Inertial Measurement Unit (IMU) was used to generate ground truth from the gyro angular rates~\cite{Rueckauer2016}.

Similarly to the tuned Gabor filter, the SNN in~\cite{Orchard2013} forms layers of neurons responding to eight speeds, eight directions, and on/off events on a 5x5 pixel region. The approach mimics the classic Lucas-Kanade in a bio-inspired framework. In contrast,~\citet{ParedesValles2020} demonstrate learning the neurons’ connection parameters from unsupervised data with a hierarchical SNN architecture. To this aim, they introduce a novel adaptive mechanism for the Leaky Integrate-and-Fire neurons and a stable implementation of the SpikeTiming-Dependent Plasticity (STDP) learning protocol. Additionally, they released the code for simulating large SNN on GPU-accelerated hardware in an open-source library, cuSNN. More recent approaches, \eg~\cite{Lee2020, ParedesValles2021, Zhang2023}, have drastically improved the accuracy performance either by combining artificial neural networks for extending to deeper layers or by adopting more complex architectures~\cite{Parameshwara2021}.

Instead of computing flow on the raw events,~\citet{Bardow2016} propose jointly estimating the image log intensity with the velocity field in a sliding window variational optimization scheme. Besides demonstrating high dynamic range frame reconstruction, this approach can obtain a dense optical flow field. However, in areas where events have not been received, the optical flow is less reliable as they result only from the constraints in the optimization equations, such as smoothing regularisation terms.

Alternative to Lucas-Kanade-inspired works,~\citet{Liu2018} propose a method based on block matching, a technique widely used for video compression. They extend their previous FPGA implementation~\cite{Liu2017} with more efficient computations for real-time operation. Remarkably, they accumulate 2D histograms of events in three adaptive time slices that are continuously rotated. Then, they found the best matching block of a region centered around an incoming event using the Sum of Absolute Difference (SAD) function. Subsequently,~\citet{Liu2022} propose a further improvement based on a novel corner detection algorithm implemented in hardware, SFAST, which allows skipping computations for non-keypoints events.

Furthermore, deep learning has been applied to leverage the large availability of data. Due to the lack of ground-truth optical flow in the event domain, initial work approached the problem following a self-supervised learning paradigm~\cite{Zhu2018, Zhu2019, Ye2020}. They adopt models, \eg U-Net, and loss functions from the standard camera deep learning research that can learn the 3D structure and the motion of the camera together with the flow. Moreover, the diverse nature of events requires finding the input representation that preserves the most information~\cite{Sun2022}. Hence, while exploring slight variations in the input format, recent methods introduce correlation cost volumes~\cite{Gehrig2021a}, recurrent units~\cite{Ding2022}, and transformer blocks~\cite{Tian2022} usually in an encoder-decoder architecture fashion. More recently, BlinkSim~\cite{Li2023}, a simulator of actual event data and optical flow ground truth based on the Blender 3D engine, has been released, allowing further tuning of deep learning models.

\subsection{Camera Localization and Mapping}
\label{sec:poses}

Estimating a camera's 6-DoF pose is fundamental for enabling autonomy in robotics and vision systems, underpinning tasks such as navigation, mapping, and interaction with the environment~\cite{bavle2023slam}. When both mapping and localization occur simultaneously in an unknown environment, the task is referred to as Simultaneous Localization and Mapping (SLAM)~\cite{taketomi2017visual,Tsintotas2022}.

Although SLAM can rely on various sensors, including LiDAR, IMU, RADAR, or even radio-based methods~\cite{kabiri2024graph}, event cameras naturally align with visual SLAM (VSLAM). Their low latency, high temporal resolution, and robustness to motion blur and lighting changes make them attractive alternatives. However, the asynchronous data stream they produce challenges conventional SLAM pipelines, which typically assume a fixed frame input.

Event cameras output sparse, high-frequency brightness changes rather than global frames. As a result, SLAM algorithms must be restructured to handle this format, often using feature-based or direct methods. Many systems represent the scene with semi-dense edge maps co-estimated with camera pose, leveraging that events are primarily triggered by edge motion~\cite{Gallego2017,Reinbacher2017,Rebecq2017}.

Early event-based SLAM systems were predominantly feature-based, extracting and tracking corners or lines to estimate motion and build sparse 3D reconstructions~\cite{Weikersdorfer2013,Weikersdorfer2014,Kim2014}. Corner detectors such as eHarris~\cite{Vasco2016}, eFAST~\cite{Mueggler2017a}, and FA-Harris~\cite{Li2019} were adapted to event data but often struggled with noise and motion variation. More recent methods improved stability by incorporating learning-based feature extractors, including recurrent networks and time-surface representations. Line-based tracking also added geometric constraints, and feature positions were typically optimized using probabilistic filters or bundle adjustment~\cite{Rebecq2017a,Kim2016}.

Direct approaches avoid explicit features and instead align event data with geometric or photometric models. Common strategies involve transforming events into image-like representations such as time surfaces, then aligning them against known scene structure or intensity maps~\cite{Gallego2018a,Kim2014}. Bayesian filtering is often used for incremental motion estimation, while methods like EVO~\cite{Rebecq2017} align event images with semi-dense maps. EMVS~\cite{Rebecq2018} introduced an efficient back-projection method to accumulate events in 3D space and recover depth from multiple viewpoints.

To improve performance in low-texture regions or during fast motion, many systems integrate IMU measurements. Visual-Inertial Odometry (VIO) pipelines such as Ultimate SLAM~\cite{Vidal2018} or ESVIO~\cite{Chen2022} fuse event and inertial data, often using continuous-time trajectory models. Stereo event cameras have also been employed to recover depth through temporal and spatial consistency~\cite{Zhou2021,Liu2023}, while RGB-D setups like DEVO~\cite{Zuo2022} combine event streams with depth sensors to enhance mapping fidelity.

Motion compensation remains key to improving spatial coherence. Techniques such as contrast maximization~\cite{Gallego2017} or event cloud alignment aim to sharpen accumulated events, supporting robust tracking even under fast motion or extreme lighting.

Loop closure and long-term consistency, while less explored, are gaining traction. Recent work applies spatiotemporal descriptors and graph-based optimization to reduce drift and improve global accuracy.

Deep learning has also become central to event-based SLAM. Early self-supervised approaches by \citet{Zhu2019} and \citet{Ye2020} showed that depth, optical flow, and ego-motion can be learned jointly from voxel-grid or time-surface representations. These models typically use CNN encoder-decoders trained with photometric or warping losses.

Subsequent work improved monocular depth estimation. \citet{HidalgoCarrio2020} used recurrent CNNs to accumulate spatiotemporal information and predict dense depth from events alone. EMoDepth~\cite{zhu2024emodepth} refined this with a cross-modal training strategy: using aligned frames only during training while operating with events alone at inference, achieving state-of-the-art accuracy on MVSEC and DSEC.

Pose relocalization also benefited from deep models. CNN–LSTM networks~\cite{Jin2020} and transformer-based approaches like AECRN~\cite{Hu2022aecrn} exploit entropy-based event representations to regress 6-DoF pose. PEPNet~\cite{Ren2024pepnet} introduced a point-based model that processes raw event streams as 4D point clouds, outperforming prior work while remaining lightweight. 
Spiking neural networks (SNNs) have been explored for their potential efficiency on neuromorphic hardware. Spike-FlowNet~\cite{Lee2020spikeflownet} combined ANN and SNN layers for optical flow, while a fully spiking approach by \citet{Hagenaars2021snnflow} achieved comparable results with much lower energy cost.
Although many learned methods still rely on auxiliary frames or depth maps during training, the trend is moving toward fully event-driven models. Progress in spatiotemporal event representations—such as entropy frames, voxel grids, or point clouds—alongside attention modules, recurrent encoders, and spiking networks, is making real-time, frame-free SLAM increasingly feasible.

For evaluation, most approaches rely on public benchmarks such as MVSEC~\cite{Zhu2018a}, DSEC~\cite{Gehrig2021b}, the IJRR Event Camera Dataset~\cite{Mueggler2017}, and M3ED~\cite{Li2023m3ed}.

Event-based SLAM remains an evolving frontier. While feature-based and direct methods offer complementary strengths, major challenges persist in scalability, robustness, and fusion. Continued development of hybrid pipelines, neuromorphic hardware, and self-supervised learning is likely to drive future advances in autonomous event-based systems.

\subsection{Moving Object Detection}

Motion detection clearly highlights the advantages of neuromorphic photoreceptors compared to standard cameras. Thanks to their event-driven nature, neuromorphic sensors offer higher temporal resolution and faster responses, providing a more efficient way to detect moving objects. Unlike standard cameras, which rely on sequences of intensity frames and indirect measurements (such as optical flow), event cameras directly sense motion as changes occur in the scene. Under constant lighting and stationary camera conditions, segmenting moving objects becomes relatively straightforward, as only moving elements trigger events~\cite{Litzenberger2006a}. However, when the camera itself is moving, separating object motion from the camera’s ego-motion becomes more complex.

Initial efforts to tackle this challenge relied on classical computer vision techniques adapted to neuromorphic sensing. For instance, \citet{Glover2016} successfully tracked a fast-moving ball with an event camera mounted on the iCub robot by integrating Hough-transform circle detection with optical flow techniques, achieving robust detection at 500 Hz despite significant background clutter caused by robot movement. Similarly, \citet{Vasco2017} leveraged the joint velocities of the robot to distinguish the motion of independent objects from the motion induced by the camera, effectively tracking the general shapes of objects.

To improve robustness under ego-motion, researchers explored motion-compensated representations of event data~\cite{Gallego2018}, which align events into sharp images by estimating and removing the camera’s motion. This approach enabled \citet{Mitrokhin2018} to detect moving objects through motion inconsistencies, and was further extended by \citet{Stoffregen2019}, who introduced a clustering method that jointly estimates object motions to refine segmentation results.

As deep learning entered the field, early models were adapted specifically for object detection using event data. \citet{Cannici2019} proposed YOLE and fcYOLE, two neural architectures designed to process events either through integrated surfaces or in a fully asynchronous manner. These models demonstrated the feasibility of adapting frame-based convolutional techniques to sparse event streams. Building on these ideas, \citet{Liang2022} introduced GFA-Net and CGFA-Net—transformer-based detectors evaluated on the EventKITTI dataset that combine local feature extraction with global context through edge-aware position encoding.

Expanding on these foundational approaches, \citet{Mitrokhin2019} presented a more integrated neural-network-based pipeline for motion segmentation. Their model simultaneously estimated depth, ego-motion, segmentation masks, and object velocities. They also introduced the EV-IMO dataset, providing detailed pixel-wise annotations in challenging indoor scenes. Later, the EVIMO2 dataset~\cite{burnerEVIMO2EventCamera2022} expanded these benchmarks with greater complexity and more extensive annotations, facilitating robust training for both supervised and semi-supervised methods.

In parallel, neural architectures were refined to better exploit the asynchronous nature of events. For instance, \citet{Sekikawa2022eventcnn} demonstrated efficient convolutional neural networks utilizing event-driven time surfaces for accurate segmentation. Spiking neural networks, previously explored for optical flow tasks, have also shown promise in segmentation. SpikeMS~\cite{Parameshwara2021} applied a deep spiking encoder–decoder architecture to motion segmentation using DVS input, achieving performance comparable to artificial neural networks while significantly reducing energy consumption.

Recent approaches, like the Recurrent Vision Transformer (RVT) by~\citet{Gehrig2023rvt}, began employing transformer architectures to fully leverage event data's temporal and spatial properties. RVT reached state-of-the-art results on automotive detection benchmarks (Prophesee GEN1), achieving extremely low latency detection and demonstrating that transformer models could significantly enhance event-based object detection.

To address complex outdoor scenes where ego-motion plays a dominant role, methods like EmoFormer by~\citet{Zhou2024emoformer} have emerged. EmoFormer cleverly uses events only during training to inject strong motion awareness into a segmentation network, which then performs segmentation using only standard images at inference. They introduced the DSEC-MOS dataset, providing pixel-wise motion annotations for driving scenarios and addressing a critical gap in available training data. A complementary approach by~\citet{Georgoulis2024outofroom}, called ``Out of the Room'', explicitly compensates for ego-motion using monocular depth estimation before segmenting independently moving objects, further setting new benchmarks on EV-IMO and DSEC-MOTS datasets.

Given the difficulty and cost of labeling event data, recent methods also explored unsupervised or semi-supervised strategies. Un-EvMoSeg by~\citet{Wang2023unev} introduced an entirely unsupervised method using geometric constraints to detect independently moving objects without needing labeled data, achieving competitive results compared to supervised approaches. Similarly, LEOD, proposed by~\citet{Wu2024leod}, uses pseudo-labels and temporal consistency to train detectors effectively with minimal supervision, demonstrating strong results with very few annotations.

Beyond neural networks, researchers have also drawn inspiration from biology. The retina-inspired Object Motion Sensitivity (OMS) algorithm by~\citet{Bak2024oms} mimics retinal circuits, providing a lightweight and efficient way to isolate moving objects without explicit ego-motion compensation. Another non-learning approach, JSTR by~\citet{Zhou2024jstr}, combined IMU measurements and geometric reasoning to segment moving objects effectively, showcasing robust results without relying on heavy learning frameworks.

Hybrid methods combining event data with other modalities, particularly RGB frames, have also proved valuable. For instance, RENet~\cite{Zhou2023renet} fuses event data and standard RGB images using attention mechanisms, greatly improving object detection accuracy under diverse conditions, including challenging lighting and rapid motion scenarios. Another notable hybrid approach, FlexEvent, introduced by~\citet{lu2024flexeventeventcameraobject}, focuses on adapting object detection to arbitrary event frequencies. It combines event data with RGB frames using an adaptive fusion module (FlexFuser) and a frequency-adaptive learning strategy (FAL), achieving robust object detection performance across frequencies ranging from 20Hz up to 180Hz. This flexibility makes it particularly suitable for dynamic, real-world scenarios where event rates vary significantly.

As new datasets expand the range of evaluation scenarios, the field steadily bridges the gap between low-level motion cues and high-level scene understanding. With approaches ranging from fully event-based models to hybrid and unsupervised methods, current systems are increasingly capable of accurate, real-time segmentation, even in challenging, dynamic environments.

\section{Applications}
This section reviews several applications that may benefit from introducing event cameras and image-processing algorithms. Event cameras are specialized types of sensors that can be used in various applications. Their unique features, such as high temporal resolution, low latency, and low power consumption, make them particularly useful for applications that require real-time processing and low latency.

The principal qualities of event cameras, namely the high temporal resolution (up to 1 microsecond), the low latency deriving from independent pixel chips, low power consumption (around 1mW), low memory footprint derived from the sparse output, and a high dynamic range, which enables vision with direct sunlight or illumination comparable to moonlight~\cite{Sun2021}, reveal their potential in such scenarios where the tasks are susceptible to the requirement of robustness, safety, and accuracy in demanding environmental conditions. For example, several civil applications, such as search and rescue or surveillance, autonomous driving~\cite{chen2020event}, traffic monitoring~\cite{Litzenberger2006}, power line inspections~\cite{Dietsche2021}, industry 4.0, star tracking~\cite{Chin2019} and space situational awareness~\cite{Cohen2019, Afshar2020} may profit from this vision platform earlier than commercial products, as their impact on society attracts more funds. On this note, Augmented or Virtual Reality (AR/VR), which would benefit from their characteristics for accurate device motion estimation, gesture recognition, or eye tracking~\cite{Angelopoulos2021} with low latency, will probably attract once the NC reaches the mass markets. 
 
Herein, we describe some practical tasks for which the unique characteristics of event cameras motivate their adoption in civil applications : 

\subsection{Health and Sport-activity Monitoring}
Event cameras can capture detailed information about an individual's activity, including detecting falls, tracking movements, and analyzing gait patterns, and could be applied to provide early warning signs of health issues or injuries. Several works have recently been proposed to estimate the human body pose from event camera measurements, \eg ~\cite{Colonnier2021, Zou2021, Chen2022, Scarpellini2021, Shao2023, Zhang2023a}, for which task a dedicated dataset has been released~\cite{Calabrese_2019_CVPR_Workshops}. Furthermore, combining the high temporal resolution events with color images allows for interpolating new frames faster than the original video stream, reducing ghosting and other artifacts caused by non-linear motions~\cite{Tulyakov2021, Tulyakov2022}.

\subsection{Industrial Process Monitoring}
Event cameras are emerging as powerful tools for industrial environments that demand high-speed, high-precision monitoring. Their low latency, high temporal resolution, and robustness to lighting variations make them particularly well-suited for real-time quality control, equipment diagnostics, and predictive maintenance.

One illustrative use case is high-speed object counting. For example, \citet{Bialik2022} demonstrated a Prophesee EVK1 event camera successfully counting corn grains on a fast-moving feeder line, showcasing the potential of neuromorphic vision in manufacturing and logistics applications. 
Beyond object counting, NCs have shown promise in broader industrial process monitoring tasks. For instance, \citet{dold2025event} investigated the use of event cameras for laser welding, a domain where conventional photodiodes and high-speed cameras are typically used. Their study demonstrated that event cameras could visualize welding dynamics with superior temporal fidelity and detect production anomalies using learned representations.
In vibration monitoring, a critical task for predictive maintenance and structural diagnostics, \citet{baldini2024measuring} used event cameras to track mechanical vibrations with an accuracy comparable to expensive laser Doppler vibrometers. Their system combined stereo event tracking and video reconstruction (via E2VID~\cite{Rebecq19cvpr,Rebecq19pami}) to measure subtle displacement patterns at sub-pixel resolution.

These examples reflect the increasing adoption of event cameras for industrial process monitoring, which requires high-frequency observation and fast decision-making from precision manufacturing to large-scale industrial systems.

\subsection{Space Sector} Neuromorphic sensors can be applied to telescopes to track stars~\cite{Chin2019}, satellites~\cite{bacon2021satellite}, or debris in orbit~\cite{zolnowski2019observational} from the ground to avoid potential damage to other infrastructures. Recent research suggests that NCs with high spatial and temporal resolution may be exploited to identify the material of satellites~\cite{jolley2019use}. Furthermore,~\citet{Jawaid2022} leverage the high dynamic range of the event sensor to estimate satellite pose to ensure robustness to drastic illumination changes. Also, \citet{Mahlknecht2022} demonstrate that event cameras are suitable for planetary explorations where challenging scenarios, such as the Mars landscape, pose many challenges in estimating the autonomous robot self-position. The International Space Station (ISS) has an event-based sensor to detect lightning and sprite events in the mesosphere. These events can occur in as little as 100 microseconds~\cite{McHarg2022}.

\subsection{Surveillance \& Search and Rescue}
NCs are well suited for monitoring public spaces or securing buildings due to their low power consumption and real-time processing capabilities. These features enable them to monitor large areas without frequent maintenance or battery replacements. Furthermore, traditional surveillance cameras can be limited in accurately detecting and tracking objects in complex and dynamic environments where the HDR capabilities of these sensors compensate for lighting variability factors and nighttime operations. Research has shown that NCs can detect and track multiple moving objects in real-time, even if many challenges of complex environments must still be addressed~\cite{Hinz2017}.~\citet{Ganan2022} propose an event-based processing scheme for efficient intrusion detection and tracking of people, using a probabilistic distribution and CNN, which has been validated in various scenarios on a DJI F450 drone.  

Aerial robots are the primary platform for surveillance and search and rescue applications. Mainly, event-based cameras' high temporal resolution and dynamic range help handle the motion blur caused by UAVs while detecting and tracking possible intruders~\cite{Dios2020}. Recent work by~\citet{rodriguez-gomezAsynchronousEventbasedClustering2020} introduces an asynchronous event-based clustering and tracking method for intrusion monitoring in UAS. Their approach leverages efficient event clustering and feature tracking while incorporating a sampling mechanism to adapt to hardware constraints, demonstrating improved accuracy and robustness in real-world scenarios. Deep learning methods for event-based human intrusion detection in UAV surveillance have also been explored to gain more confidence in determining the type of moving object. \citet{perez-cutinoEventbasedHumanIntrusion2021} present a fully event-based processing scheme that detects intrusions as clusters of events and classifies them using a CNN to determine whether they correspond to a person. In particular, this method eliminates the need for additional onboard sensors and fully exploits the asynchronous nature of event cameras. 

Similarly to surveillance, event cameras are useful in search and rescue operations, especially in environments like forests or mountains, because of their high temporal resolution and low power consumption, and allow them to capture detailed information about the environment and provide real-time feedback to rescuers while moving fast and for extended missions. More importantly, the low latency of NCs can allow remote UAV pilots to perform more aggressive flights~\cite{Falanga2019}, which is critical to reducing operation time while safely avoiding obstacles in cluttered unknown environments~\cite{Loquercio2021}. 

\subsection{Autonomous Driving} Autonomous driving applications like collision avoidance could benefit from event cameras. With their low latency and high temporal resolution, they can capture detailed information about the environment and provide real-time feedback to the autonomous driving system. Hence, they will be an essential resource for implementing advanced driver assistance systems (ADAS) and self-driving cars in the future. To this aim, \citet{Wzorek2022} recently demonstrated how a neural network could detect traffic signs. Not only, but event cameras have also been tested on the driver distraction detection task by~\citet{Yang2022}, where the authors evaluated the proposed approach by converting standard video clips with an event simulation tool~\cite{Hu2021-v2e-cvpr-workshop-eventvision2021}.

\subsection{Traffic Monitoring} Event cameras may be helpful for traffic monitoring applications, such as estimating car speed~\cite{Litzenberger2006}. Their low power consumption and real-time processing capabilities make them well-suited for monitoring large areas without frequent maintenance or battery replacements. Therefore, event cameras can detect and track multiple cars on the road simultaneously~\cite{Shair2022} or pedestrians and cyclists~\cite{Belbachir2010}.

\subsection{Defense}
Event cameras offer significant advantages for defense applications due to their low power consumption, high temporal resolution, and ultra-low latency. These features make neuromorphic cameras ideal for embedded systems in UAVs and other autonomous platforms, enhancing obstacle detection, target tracking, and surveillance while maintaining power efficiency. In reconnaissance and battlefield monitoring, NCs provide continuous high-speed data streams that improve situational awareness in real time~\cite{kirkland2020neuromorphic}. Their ability to track fast-moving targets is particularly valuable for Unmanned Ground and Underwater Vehicles (UGVs and UUVs), where reaction time is critical~\cite{stewart2021drone}.

Recent studies have also explored using NCs for laser warning and detect-before-launch (DBL) capabilities. For instance, \citet{boehrerLaserWarningPointed2024a} demonstrate how the high temporal resolution of event cameras can be leveraged to detect laser emissions and retro-reflections from pointed optics key indicators of hostile intent. Their system was evaluated in operational scenarios during the DEBELA trial, showing that event-based sensing enables early and reliable threat detection. Complementing this work, the DEBELA project~\cite{eiseleDEBELAInvestigationsPotential2024} investigates electro-optical technologies for future self-protection systems, focusing on within-visual-range missile threats that are difficult to detect using conventional sensors.

NCs also show promise in Counter-Unmanned Aerial Systems (C-UAS), where their ability to capture fast-moving drones or hypersonic missiles can aid early warning systems. Their sensitivity in the infrared and shortwave infrared bands allows for enhanced night vision and detection of low-signature propellants~\cite{kim2019neuromorphic,cha2014neuromorphic}. Together, these capabilities position event-based sensors as powerful tools for modern defense, offering real-time threat perception, reduced false alarms, and greater autonomy in decision-making.

\subsection{Others}
The range of industries exploiting neuromorphic cameras is unlimited. In agriculture, their asynchronous operation and high temporal resolution open new avenues for real-time crop monitoring and precise field management, enhancing precision farming techniques~\cite{mavridou2019machine}. In healthcare, event cameras are being explored for applications such as surgical monitoring and neural imaging, where the ability to capture subtle, fast physiological motions can improve diagnostic accuracy~\cite{moeys2017sensitive,choi2024inspiration}.

Moreover, neuromorphic vision systems can be integrated into sensor fusion frameworks, combining modalities such as inertial sensors, microphones, or bio-signals to enhance situational awareness. For instance, \citet{Kiselev2016} demonstrate a real-time FPGA-based system combining a DVS with a Dynamic Audio Sensor (DAS), achieving significantly higher classification accuracy through multi-modal input. Similarly, \citet{OtextquotesingleConnor2013} present a spiking Deep Belief Network that fuses input from a silicon retina and cochlea, achieving robust performance even under sensory noise. These examples highlight how event-based fusion can enrich perceptual systems in fields like mobile robotics, smart wearables, and embedded AI.

\section{Discussion}

\subsection{Summary}
Neuromorphic vision sensors, or event-based cameras, represent a fundamental shift from traditional frame-based imaging. Unlike conventional cameras that record full images at fixed intervals, neuromorphic cameras capture visual information asynchronously by registering changes in brightness at each pixel. This results in sparse, low-latency, and highly efficient data regarding power consumption and storage. In addition, neuromorphic sensors have exceptionally high temporal resolution and dynamic range, allowing them to operate effectively in challenging lighting conditions and rapidly changing environments.

This review systematically covered the key dimensions of neuromorphic vision technology: the evolution of sensor hardware, the specialized algorithms developed to process event-based data, and their diverse applications. Hardware advancements highlighted include sensor architectures that evolved from early silicon-retina concepts to increasingly sophisticated designs that capture richer visual information, including colors and absolute light intensity, at higher resolution. Algorithmically, event-based processing has adapted and extended classical image-processing tasks, \eg feature detection, optical flow estimation, visual odometry, and object tracking, to handle asynchronous event streams efficiently. Finally, neuromorphic cameras have demonstrated substantial potential in various practical fields, including robotics, autonomous vehicles, industrial automation, and surveillance, taking advantage of their unique capabilities to enhance real-time responsiveness and robustness to environmental dynamics.

In the following sections, we dive deeper into analyzing the identified gaps of the current stage of development in the field of neuromorphic vision and draw some insights into the direction research and industry could take to capture the multiple opportunities this sensor offers. Hence, we provide an overview of the conclusion we derived in \autoref{table:discussion}.

\begin{table*}[!htpb]
\centering
\begin{tabular}{ >{\centering}p{1.5cm} >{\centering}p{5.8cm} >{\centering\arraybackslash}p{6.8cm} }
\Xhline{3\arrayrulewidth}
\textbf{Level} & \textbf{Gap Analysis} & \textbf{Future Directions} \\
\Xhline{3\arrayrulewidth}
\rowcolor{lightgray} 
\multirow{5}{1.5cm}
& Sensor availability & Lower manufacturing costs, mass production \\
& Manufacturing complexity & Industrial collaborations (\eg, Prophesee) \\ 
\rowcolor{lightgray} 
\centering \textbf{Hardware} & High sensor cost & Infrared, non-visible spectrum \\
& Low spatial resolution & Neuromorphic chips for consumer electronics \\
\rowcolor{lightgray} 
& Limited spectral range & Edge devices, low power, real-time processing \\
\hline
\multirow{5}{1.5cm}& Immature event-based algorithms & Improve SNNs, LSTMs \\
\rowcolor{lightgray}
  & Lack of universal data representation & Neural Event Stacks (NEST) \\
\textbf{Algorithmic} & Real-time processing challenges & Transformer models (object detection, \etc) \\
\rowcolor{lightgray} 
& Lack of benchmarks & Standardized event-based benchmarks \\
& Sparse data management & Synthetic event-data generation (\eg v2e) \\
\hline
\rowcolor{lightgray}
\multirow{5}{1.5cm}& Lab prototypes & Real-world industrial solutions \\
& Poor performance in dynamic environments & Integrate with traditional sensors \\
\rowcolor{lightgray} 
\textbf{Applications} & Poor event-based SLAM performance & Event-based SLAM in complex environments \\ 
& Limited commercial solutions & Autonomous driving, robotics, surveillance \\
\rowcolor{lightgray} 
& Drift, loop closure issues & Low latency, high temporal resolution \\
\Xhline{3\arrayrulewidth}
\end{tabular}
\vspace{5pt} 
\caption{Gap analysis and Future directions overview.}
\label{table:discussion}
\end{table*}

\subsection{Gaps Analysis}
Despite considerable advancements in neuromorphic sensors and algorithms, several gaps remain that prevent widespread adoption and limit their full replacement of classical vision sensors.

At the hardware level, the primary limitations are sensor availability, manufacturing complexity, and cost. Neuromorphic sensors remain expensive due to their specialized manufacturing processes, restricting broad commercial availability. Additionally, current event cameras typically provide lower spatial resolution compared to traditional frame-based sensors, limiting their effectiveness in applications demanding high detail. Another significant hardware constraint is the limited spectral range, with most sensors operating only in the visible spectrum. Although early initiatives like the DARPA FENCE program and recent developments of infrared-sensitive neuromorphic sensors (\eg SWIR-sensitive cameras) exist, these efforts are still at an early stage, limiting widespread implementation.

Algorithmically, a major challenge arises from fundamental differences between event-based and conventional visual data. Event-based vision algorithms are comparatively less mature and require new data representation methods and processing approaches. Although methods like voxel grids, time surfaces, and event histograms have emerged, a universally accepted approach adaptable across multiple vision tasks is still lacking. The continuous and asynchronous nature of sparse event streams poses significant challenges for developing robust algorithms and represents a substantial paradigm shift from traditional computer vision techniques. Notwithstanding the sparse nature, real-time processing and intelligent clustering of event streams remain challenging, as managing the high volume of events and extracting relevant information is nontrivial.
Additionally, benchmarks and standardized evaluation frameworks designed explicitly for event-based data remain limited, impeding progress in algorithm validation.

At the application level, neuromorphic vision systems are primarily limited to laboratory prototypes, with few robust, commercially viable solutions available. Achieving consistent performance in uncontrolled, dynamic environments remains difficult, particularly significant challenges arise from environmental noise such as intermittent lighting variations and sensor-induced noise, requiring more advanced noise filtering methods. For example, critical tasks such as event-based visual SLAM that is paramount in the future of autonomous driving or other robotic context, still struggle with drift reduction, effective loop closure detection, and reliable operation in complex real-world scenarios involving rapid movements or significant scene aspect changes.

Finally, fully exploiting the inherent energy-efficiency advantages of neuromorphic sensors in practical deployments demands integration with specialized neuromorphic computing hardware, optimized explicitly for processing sparse and asynchronous event data. Current general-purpose hardware, such as CPUs and GPUs, lack the efficiency for event-based processing, while dedicated neuromorphic computing platforms that support SNNs remain limited in commercial availability. Achieving widespread industrial use will require concerted efforts toward hardware innovation and software maturity, a challenge most companies are currently unable to tackle without greater standardization and market maturity.

\subsection{Opportunities and Future Directions}

However, despite the gaps discussed, several promising opportunities exist for further advancing neuromorphic vision technology.

In hardware, key opportunities include reducing sensor manufacturing costs through mass production and strategic industrial collaborations. Recent partnerships, such as that between Prophesee and Qualcomm, which aim at integrating event-based cameras into smartphones, and Google's integration of neuromorphic sensors into Android XR for augmented reality, are paving the way for broader market adoption. Furthermore, neuromorphic chips like SynSense Speck, designed for ultra-low-power and high-speed imaging, can potentially extend event-based sensing to consumer electronics and affordable machine vision solutions. Expanding into infrared and non-visible spectral domains also presents significant potential, particularly for security, defense, and environmental monitoring applications. Additionally, integrating neuromorphic sensors into edge devices paired with neuromorphic processors, which significantly reduce power consumption and enhance real-time processing capabilities, presents another critical opportunity for practical implementations, especially in energy-constrained environments.

On the event-processing side, improving temporal neural networks remains important, including architectures like Spiking Neural Networks (SNNs) and Long Short-Term Memory (LSTM), which naturally handle the time-based event data. A growing body of work also explores learned event-based representations, which encode spatiotemporal patterns in formats better suited to downstream processing, signaling space for improvement in this area. Moreover, transformer-based models, initially developed for language processing~\cite{Vaswani2017} and later adapted to traditional computer vision~\cite{Lv2023}, are starting to show potential for event-based vision tasks such as object detection, video reconstruction, and pose estimation. These models effectively capture long-term temporal dependencies in event data, offering advantages over conventional convolutional networks. In this context, sparse-aware transformer designs like the Event Transformer (EvT)~\cite{Sabater2022} further improve computational efficiency by leveraging the unique sparsity of event streams, making them more suitable for real-time, resource-constrained applications.

Developing standardized deep learning benchmarks and datasets specifically for event-based vision tasks is critical to accelerating algorithmic maturity and adoption~\cite{zheng2023deep}. Advances in synthetic event-data generation tools (\eg v2e~\cite{Hu2021-v2e-cvpr-workshop-eventvision2021}) that accurately emulate sensor behavior under varying conditions also offer significant potential to facilitate algorithm development and training, reducing dependency on extensive real-world data collection. These tools can further enhance algorithm robustness to environmental factors, such as noise, varying illumination, and complex scenes, by providing an extensive and controllable source of training data for neural network-based methods. Lastly, developing computationally efficient algorithms optimized for specialized neuromorphic hardware accelerators remains essential for enabling practical and widespread adoption.

Regarding applications, moving from laboratory prototypes to real-world industrial solutions remains a significant opportunity. Integrating neuromorphic sensors with traditional cameras and other sensor types (such as IMUs, LiDAR, and microphones) can combine strengths and significantly enhance system performance. In particular, event-based SLAM systems that leverage both neuromorphic sensing and neuromorphic computing represent an immediate opportunity, especially in complex environments where conventional sensors struggle, such as autonomous vehicles navigating dynamic urban settings, drones operating under variable lighting conditions, or robotic systems employed in search-and-rescue and defense applications. 

Furthermore, several less-explored application domains could notably benefit from neuromorphic sensors, opening new opportunities for adoption. For instance, agriculture and precision farming can leverage event-based vision, \eg for real-time crop monitoring. Healthcare applications, particularly surgical assistance, patient monitoring, or even microexpression analysis for telemedicine, could exploit the sensitivity of neuromorphic sensors to rapid and subtle physiological changes. Additionally, applications in sports analytics, such as real-time ball tracking or athlete movement analysis, present another promising use case, given the sensor's ability to precisely track high-speed objects without motion blur. Even in heavy industry, where traditional high-speed cameras are already used for equipment inspection and wear monitoring~\cite{kingdon2015eyes}, a transition to event-based vision could improve temporal resolution and data efficiency under harsh, dynamic conditions.

Increased awareness and dissemination efforts are crucial to facilitating industrial adoption. Initiatives like the 4th International Workshop on Event-based Vision at CVPR 2025 and the NeVi 2024 Workshop at ECCV 2024 are already helping to connect academia and industry by highlighting practical benefits and driving interest in neuromorphic sensors. Similarly, industry-focused events such as the VISION Fair provide valuable opportunities to reach broader industrial stakeholders. Expanding participation in these events, supported by targeted promotional activities and strategic partnerships, will further encourage market adoption and raise industry awareness of event-based vision technologies.

In conclusion, neuromorphic vision technology is approaching a critical turning point. Clear opportunities exist to address current gaps in sensor affordability, algorithm effectiveness, and practical applications. Progress in these areas will accelerate the transition of neuromorphic vision from research novelty to widely adopted technology.


\bibliographystyle{IEEEtranN}
\bibliography{references}

\end{document}